\documentclass[conference]{IEEEtran}
\IEEEoverridecommandlockouts
\usepackage{cite}
\usepackage{amsmath,amssymb,amsfonts}
\usepackage{algorithmic}
\usepackage{graphicx}
\usepackage{textcomp}
\usepackage{xcolor}
\usepackage{multirow}
\usepackage{comment}
\usepackage{pifont}
\usepackage{url}
\usepackage{balance}

\def\BibTeX{{\rm B\kern-.05em{\sc i\kern-.025em b}\kern-.08em
    T\kern-.1667em\lower.7ex\hbox{E}\kern-.125emX}}
\begin{document}

\title{TSalV360: A Method and Dataset for Text-driven Saliency Detection in 360-Degrees Videos*\\
\thanks{This work was supported by the EU Horizon Europe programme under grant agreement 101070109 TransMIXR. \\ IEEE CBMI 2025. \copyright IEEE. This is the authors' accepted version. The final publication is available at https://ieeexplore.ieee.org/}
}

\author{\IEEEauthorblockN{Ioannis Kontostathis}
\IEEEauthorblockA{
\textit{ITI, CERTH}\\
Thessaloniki, Greece \\
ioankont@iti.gr}
\and
\IEEEauthorblockN{Evlampios Apostolidis}
\IEEEauthorblockA{
\textit{ITI, CERTH}\\
Thessaloniki, Greece \\
apostolid@iti.gr}
\and
\IEEEauthorblockN{Vasileios Mezaris}
\IEEEauthorblockA{
\textit{ITI, CERTH}\\
Thessaloniki, Greece \\
bmezaris@iti.gr}
}

\maketitle

\begin{abstract}
In this paper, we deal with the task of text-driven saliency detection in $360^{\circ}$ videos. For this, we introduce the TSV360 dataset which includes 16,000 triplets of ERP frames, textual descriptions of salient objects/events in these frames, and the associated ground-truth saliency maps. Following, we extend and adapt a SOTA visual-based approach for $360^{\circ}$ video saliency detection, and develop the TSalV360 method that takes into account a user-provided text description of the desired objects and/or events. This method leverages a SOTA vision-language model for data representation and integrates a similarity estimation module and a viewport spatio-temporal cross-attention mechanism, to discover dependencies between the different data modalities. Quantitative and qualitative evaluations using the TSV360 dataset, showed the competitiveness of TSalV360 compared to a SOTA visual-based approach and documented its competency to perform customized text-driven saliency detection in $360^\circ$ videos.
\end{abstract}


\section{Introduction}

Over the last years, there is an ongoing interest in offering a more comprehensive and immersive viewing experience to the users. In terms of content, this is supported by producing $360^{\circ}$ videos that can be consumed primarily using VR headsets. To facilitate viewers' navigation through the unlimited field of view of the $360^{\circ}$ video, several methods have been described in the literature. Some of them navigate the viewer by controlling the camera's position and field of view and defining an optimal camera trajectory \cite{su2016activity,8099633,Hu_2017_CVPR,9072511,10.1145/3306346.3323046,9284734}, while others produce a shorter version of the full-length video by performing $360^{\circ}$ video highlight detection \cite{10.5555/3504035.3504957} and summarization \cite{8578251,10.1007/978-3-031-53302-0_15}.

The first processing step of the methods above is to identify which parts of the $360^{\circ}$ video attract the viewers' attention, a task that is typically tackled by algorithms for $360^{\circ}$ video saliency detection. However, existing approaches \cite{10.1145/3240508.3240669,8578252,8551523,9072511,10.1007/978-3-030-01234-2_30,8418756,Atsal_2020,Sstsal_2022_CG,salvit360,10471541,10.1007/978-3-031-19833-5_25,10.1109/TCSVT.2024.3407685,10765391,10224292} aim to spot all salient objects/events in the $360^{\circ}$ video, and thus are not tailored to focus on specific salient objects/events of particular interest for the viewer. Hence, they cannot assist customized $360^{\circ}$ video navigation or summarization according to the users' needs. This task requires saliency detection methods that can incorporate the users' demands expressed, e.g., using a textual description of the desired salient objects/events in the $360^{\circ}$ video. To our best knowledge, existing text-driven saliency detection methods of the literature are compatible only with still images \cite{10.1007/978-3-031-78186-5_2,10689368}. 

To assist research on text-driven saliency detection in $360^{\circ}$ videos, in this paper we introduce a new dataset, called TSV360, consisting of approx. 16,000 triples of EquiRectangular Projection (ERP) frames, textual descriptions and ground-truth saliency maps, from 160 $360^\circ$ videos with diverse visual content. Then, building on the visual-based SOTA SalViT360 approach \cite{salvit360}, we develop a new method for text-driven saliency detection in $360^\circ$ videos, that incorporates a similarity estimation module and a viewport spatio-temporal cross-attention mechanism to estimate and model dependencies between different data modalities, and performs saliency detection conditioned on the input text. As a note, methods for visual object tracking and segmentation in $360^{\circ}$ videos (e.g., \cite{11090163}) could be also taken into account; however, such methods aim to identify and localize specific, predefined objects within the video. On the contrary, saliency detection methods aim to spot the most visually conspicuous regions in a video, thus being less restrictive to specific objects/events. Finally, using the TSalV360 method and the TVS360 dataset, we conduct a series of experimental evaluations and ablations to document the contribution of the introduced changes in SalViT360 and assess the capacity of TSalV360 to support text-driven saliency detection in $360^\circ$ videos. Our contributions are as follows:
\begin{itemize}
    \item We introduce the TSV360 dataset for text-driven saliency detection in $360^{\circ}$ video, which includes 16,000 triplets of ERP frames, textual descriptions and ground-truth saliency maps, enabling the training and objective evaluation of text-driven $360^{\circ}$ video saliency detection methods.
    \item We build the TSalV360 method for text-driven saliency detection in $360^{\circ}$ video, which integrates a similarity estimation module and a viewport spatio-temporal cross-attention mechanism to estimate and model dependencies between the different data modalities, and enable the generation of saliency maps conditioned on the input text. 
    \item We perform a set of evaluations using TSalV360 and the TSV360 dataset, documenting the capacity of TSalV360 to perform saliency detection based on the users' needs, and forming the basis for future comparisons in the field of text-driven 360$^\circ$ video saliency detection.
\end{itemize}

\section{Related Work}

\subsection{Text-driven saliency detection in still images}

Although the use of text has been extensively studied in tasks such as text-driven image captioning \cite{10143179,9578069}, object detection \cite{10377126,10030907}, video question answering \cite{NEURIPS2022_e726197f,10350729}, and summarization \cite{10222138,sd_vsum2025}, its use for guiding saliency detection in still images has not been investigated to a large extent thus far. Zhang et al. \cite{10.1007/978-3-031-78186-5_2} proposed the use of a multi-head fusion module that tries to explore the latent saliency correlation between visual and text modalities, to guide the image denoising process and progressively refine the generated saliency map to make it semantically relevant to the text. Sun et al. \cite{10689368} 
described an encoder-decoder network architecture which learns different representations of the visual features during the encoding process, and progressively fuses them with the text features using global and local cross-attention mechanisms during the decoding process, to get the final prediction results. Differently to these works, our TSalV360 method deals with the analysis of ERP frames $360^\circ$ video, conditioned to a textual description of the user's needs. 

\begin{figure*}[t]
\centering
\includegraphics[width=0.91\textwidth]{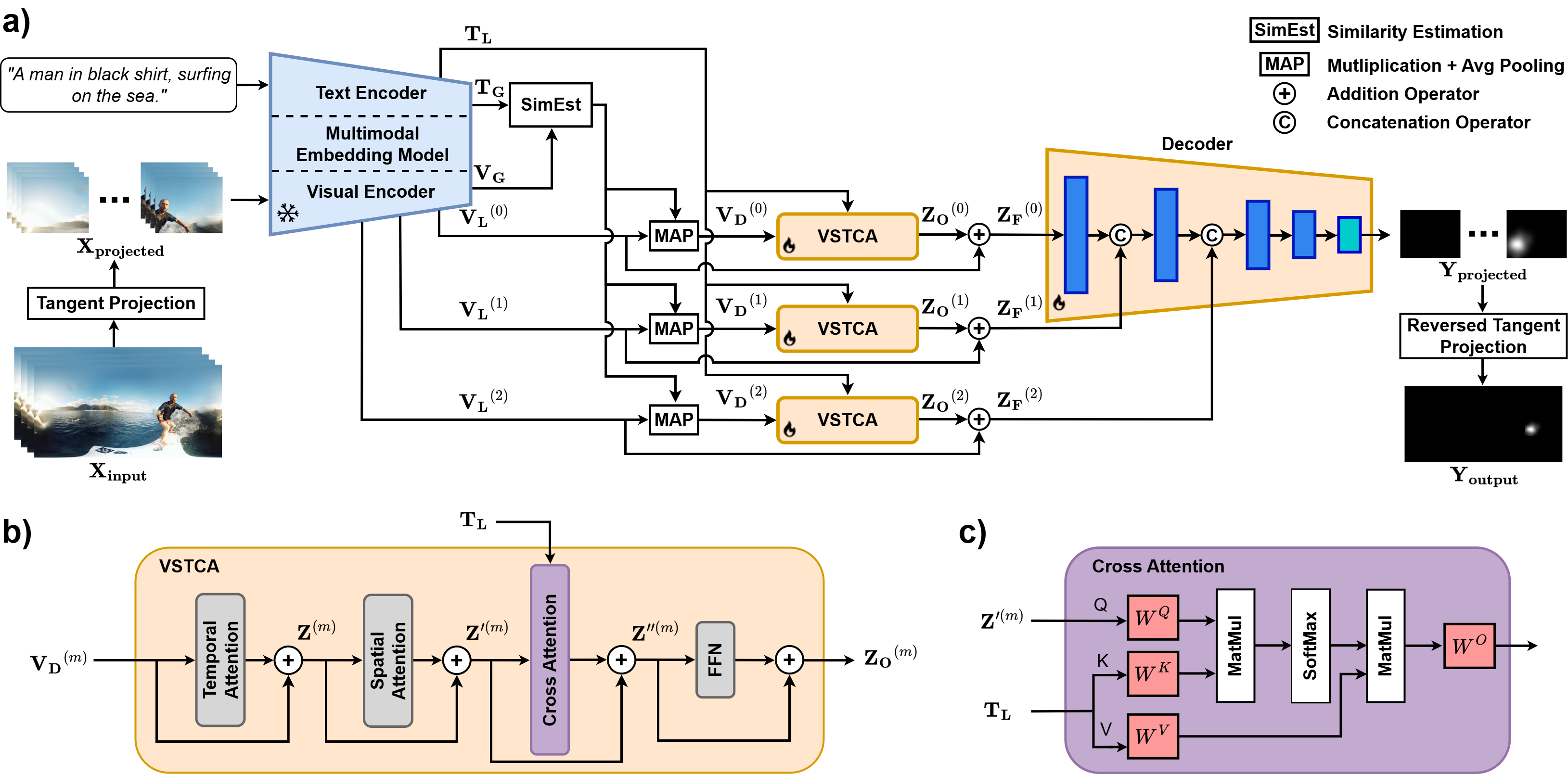}
\caption{An overview of the proposed TSalV360 method \textbf{(a)}, along with detailed presentations of the introduced VSTCA mechanism \textbf{(b)} and the implemented cross-attention \textbf{(c)} within it.} 
\label{fig1}
\end{figure*}

\subsection{Video saliency detection}

Existing methods for conventional 2D video saliency detection, that rely either on the analysis of the visual content (e.g. \cite{9706720,9010060,10045751,CHEN2021107615}), or combine different data modalities (e.g. \cite{10.1109/IROS51168.2021.9635989,LI2025113820,10655826,yu2025,10.1145/3696409.3700196,10.1109/TIP.2020.2966082}) are not applicable on $360^\circ$ videos, since they are not tailored to analyze ERP frames. For this specific type of videos, there is a different family of approaches in the literature. Nguyen et al. \cite{10.1145/3240508.3240669} fine-tuned the PanoSalNet 2D static model on $360^\circ$ video datasets to predict the saliency map of each frame without considering the temporal dimension. Cheng et al. \cite{8578252} proposed a DNN-based spatiotemporal network, comprising a static model and a ConvLSTM module to adjust the outputs of the static model according to temporal features. Qiao et al. \cite{9072511} proposed a multi-task deep neural network for head movement prediction; the center of each viewport is spatio-temporally aligned with 8 shared convolution layers to predict saliency features. Zhang et al. \cite{10.1007/978-3-030-01234-2_30} trained a spherical U-NET through teacher forcing, to apply a planar saliency CNN to a 3D geometry space. Xu et al. \cite{8418756} employed reinforcement learning to predict heatmap positions by maximizing the reward of imitating human heatmap scanpaths through the agent's actions. Dahou et al. \cite{Atsal_2020} presented a network architecture that encodes the visual features of each ERP frame using an attention model and extracts the temporal characteristics of the $360^\circ$ video using cubemap projection frames. Bernal-Berdun et al. \cite{Sstsal_2022_CG} used a Spherical ConvLST-based encoder–decoder; the encoder extracts spatio-temporal features from ERP frame sequence and the decoder leverages the latent information to predict a sequence of saliency maps. Cokelek et al. \cite{salvit360} presented the SalViT360 method which employs tangent image representations and integrates a spherical geometry-aware spatiotemporal self-attention mechanism, and trained it using a auxiliary consistency-based regularization term to reduce artifacts after inverse projection. Yun et al. \cite{10.1007/978-3-031-19833-5_25} described the Panoramic Vision Transformer which includes a ViT-based encoder with deformable convolution to enable the integration of pretrained models from normal videos without additional modules or finetuning, and performs geometric approximation only once. Finally, multimodal methods for saliency prediction in $360^\circ$ videos using also the audio modality, have been presented in \cite{10765391,10224292}. Differently from the approaches above, our TSalV360 method takes as auxiliary input a textual description of the desired salient objects and events in the video, in order to focus on the relevant regions of the $360^\circ$ video and perform customized saliency detection that meets the users' needs.

\subsection{Datasets}

Previous works on text-driven saliency detection in still images have primarily relied on the Ego4D \cite{Grauman_2022_CVPR} and SJTU-TIS \cite{10689368} datasets. Ego4D \cite{Grauman_2022_CVPR} includes 9,655 egocentric videos with daily-life activities from different scenarios (household, outdoor, workplace, etc.), that have been labeled with gaze point annotations and corresponding text descriptions at the frame level. SJTU-TIS \cite{10689368} contains 1,200 text-image pairs and the corresponding saliency maps. These text-image pairs were formulated by using 600 images with diverse visual content from MSCOCO \cite{10.1007/978-3-319-10602-1_48} and Flickr30k \cite{7410660}, and associating half of them with one and the other half of them with three manually-produced text descriptions. Saliency maps were obtained through a subjective experiment to record the eye movement data for each text-image pair. Training and evaluation of methods for saliency detection in $360^{\circ}$ videos, have been mostly done on the Pano2Vid \cite{su2016activity}, Sports-360 \cite{10.1007/978-3-030-01234-2_30}, PVS-HM \cite{8418756} and VR-EyeTracking \cite{8578657} datasets. Pano2Vid \cite{su2016activity} includes $86$ $360^{\circ}$ YouTube videos collected using specific keywords such as ``Mountain Climbing'' and ``Soccer'', while a subset of them ($20$ in total) has been annotated with human-edited NFOV camera trajectories (two per video). Sports-360 \cite{10.1007/978-3-030-01234-2_30} contains $104$ $360^{\circ}$ YouTube videos showing five sports activities (basketball, parkour, BMX, skateboarding, dance), annotated using eye fixations recorded from $27$ participants. PVS-HM \cite{8418756} comprises $76$ panoramic video sequences of diverse content (e.g., driving, sports, video games) along with data about head movement and eye fixation of $58$ humans. VR-EyeTracking \cite{8578657} consists of $208$ $360^{\circ}$ videos of various content (e.g., indoor scene, outdoor activities, music shows) that have been annotated based of eye fixation of $30$ humans.

The aforementioned datasets contain only partially the needed type of content and ground-truth annotations for supporting text-driven saliency detection in $360^{\circ}$ videos. To fill this gap in the literature, we introduce a new dataset, called TSV360, which contains 160 $360^{\circ}$ videos with visually diverse content from different topics (music shows, sports games, short movies, documentaries), coming from the VR-EyeTracking \cite{8578657} and Sports-360 \cite{10.1007/978-3-030-01234-2_30} datasets. In contrast with the existing datasets, TSV360 provides triples of ERP frames, ground-truth saliency maps and textual descriptions (approx. 16,000), thus enabling the training and evaluation of methods for text-driven $360^{\circ}$ video saliency detection.

\section{Proposed Approach}

The basis for our developments is the SalViT360 method for visual-based saliency in $360^\circ$ videos \cite{salvit360}. As described above, SalViT360 utilizes tangent image representations and models dependencies at the spatial and temporal dimension based on a spherical geometry-aware spatiotemporal self-attention mechanism. In this work, we extend SalViT360 to support text-driven saliency detection in $360^\circ$ videos, by: i) using a SOTA vision-language model that has been trained on image-text pairs for representing visual and textual input data, ii) introducing a similarity estimation module (SimEst) to estimate the semantic relevance between visual and textual features, and weight the visual local features at the output of the encoder, to force the model to pay attention to the most relevant frames to the input text, iii) replacing viewport spatio-temporal attention (VSTA) with viewport spatio-temporal cross-attention (VSTCA) that models also the dependence between visual and textual data, and iv) adding VSTCA-enhanced hierarchical skip connections between encoder and decoder, allowing the model to preserve multi-scale spatial information during decoding and the subsequent generation of the saliency map. An overview of the proposed TSalV360 method is given in Fig.~\ref{fig1}a. TSalV360 takes as input the ERP frames of the $360^\circ$ video and a textual description of the desired salient objects and events. Then, it processes the ERP frames in non-overlapping sequences of $F$ frames and predicts a saliency map for the last of them. 

\textbf{Data representation.} Each sequence of $F$ ERP frames - represented as $\mathbf{X_\text{input}} \in \mathbb{R}^{F \times C \times H_\text{in} \times W_\text{in}}$, where $C$, $H_\text{in}$ and $W_\text{in}$ denote the number of channels, height and width, respectively - is projected into $T$ tangent images \cite{tangent}, resulting in a transformed representation - denoted as $\mathbf{X_\text{projected}} \in \mathbb{R}^{F \times T \times C \times P_\text{in} \times P_\text{in}}$, where $P_\text{in}$ is the patch resolution of each tangent image. The obtained tangent images are then fed into the visual encoder of a pre-trained vision-language model to obtain a set of global visual features - indicated as $\mathbf{V_G} \in \mathbb{R}^{{F \times T \times C_G}}$, where $C_G$ is the vector's dimensionality - and a set of local visual features - denoted as $\{ \mathbf{V_L}^{(m)} \in \mathbb{R}^{F \times T \times C_m \times H_m \times W_m} \}_{m=1}^{M}$, where $M$ is the number of layers in the encoder. The textual description is given as input to the text encoder of the employed vision-language model, resulting in a set of global textual features - symbolized as $\mathbf{T_G} \in \mathbb{R}^{{1 \times C_G}}$ - and a set of local textual features - represented as $\mathbf{T_L} \in \mathbb{R}^{L_t \times C_L}$
where $L_t$ and $C_L$ stand for the length of the text tokens and the vector's dimensions, respectively. 

\textbf{Similarity Estimation module (SimEst).} The obtained global visual and textual features pass through the introduced SimEst module, which computes their cosine similarity. We consider the resulting scores as estimates about the semantic relevance between the input text and each tangent viewport across the sequence of input ERP frames, and use them to weight the visual local features at the output of the encoder ($\{ \mathbf{V_L}^{(m)} \}_{m=1}^{M}$) via element-wise multiplication. In this way, we force the model to focus on the ERP frames' spatial regions that are more semantically aligned with the input textual description. The weighted local visual features are then down-sampled using average pooling, resulting in a more condensed set of visual features - denoted as $\{ \mathbf{V_D}^{(m)} \in \mathbb{R}^{F \times T \times C_m} \}_{m=1}^{M}$. 

\begin{figure*}[t]
\centering
\includegraphics[width=0.85\textwidth]{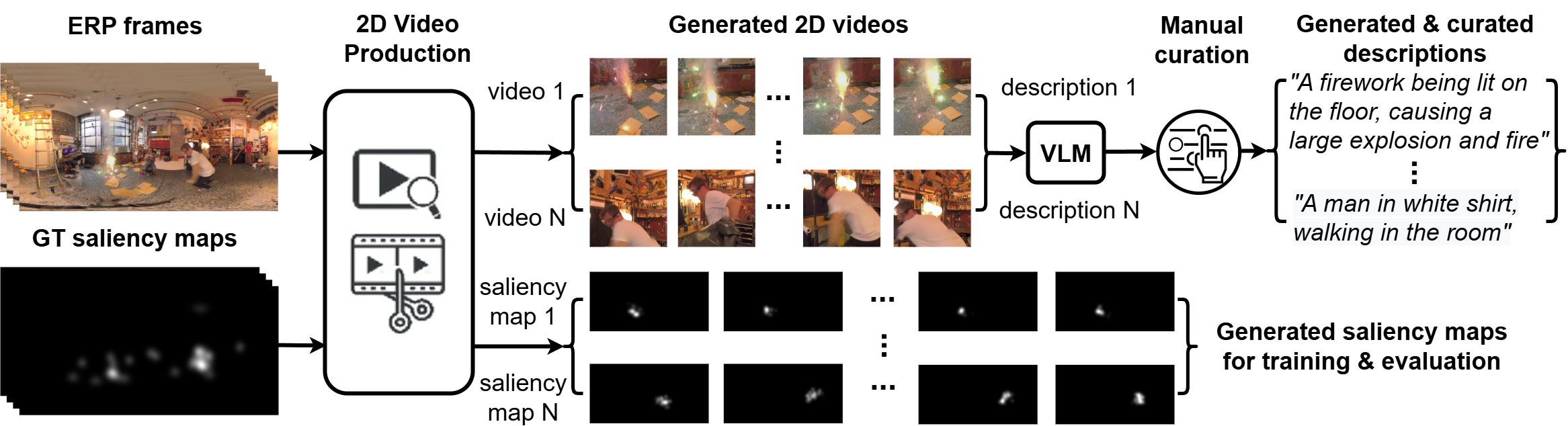}
\caption{An overview of the performed methodology for producing the TSV360 dataset.} 
\label{fig2}
\end{figure*}

\textbf{Viewport Spatio-Temporal Cross-Attention Mechanism (VSTCA).} To introduce $360^\circ$ geometry awareness into the viewport spatio-temporal attention mechanism, following \cite{salvit360}, we use learnable spherical positional embeddings for spatial information and learnable temporal embeddings to capture temporal dynamics across frames. Moreover, to model inter- and intra-frame dependencies across tangent viewports taking also into account the textual description, we introduce a cross-attention mechanism. As depicted in Fig. \ref{fig1}b, we first apply temporal attention across the same tangent viewports over the $F$ consecutive ERP frames (resulting in $\{ \mathbf{Z}^{(m)} \}_{m=1}^{M}$), and then spatial attention across the $T$ tangent viewports within each frame (obtaining $\{ \mathbf{Z'}^{(m)} \}_{m=1}^{M}$). Subsequently, we perform cross-attention, where the output of spatial attention $\{\mathbf{Z}'^{(m)}\}_{m=1}^{M}$ is used as the Query, and the textual local features $\mathbf{T_L}$ are used as Key and Value (see Fig. \ref{fig1}c). The output of cross-attention is formed as follows:
\[
\begin{aligned}
&\mathbf{Q = W^Q} \cdot {\mathbf{Z'}^{(m)}}_i , \quad \text{where } i \in [1, \dots, N] \\
&\mathbf{K, V =  W^K} \cdot \mathbf{T_L}_j, \mathbf{W^V} \cdot \mathbf{T_L}_j  \quad \text{where } j \in [1, \dots, L_t] \\
&\text{CrossAttention}(\mathbf{Q, K, V}) = \text{SoftMax}(\mathbf{QK^\top} / \sqrt{d_k}) \mathbf{V} \cdot \mathbf{W^O} \\
\end{aligned}
\]
where $N = F \times T$ is the total number of visual tokens, $\mathbf{W^Q}$, $\mathbf{W^K}$, $\mathbf{W^V}$ are the learned projection matrices for the Query, Key and Value, respectively, $\mathbf{W^O}$ is the output projection matrix, and $d_k$ is the dimension of the textual local features. The output of the cross-attention
$\{ \mathbf{Z''}^{(m)} \}_{m=1}^{M}$
 goes through a feed-forward network (FFN) and a skip connection, forming the output of the VSTCA $\{ \mathbf{Z_o}^{(m)} \}_{m=1}^{M}$. This output is fused with the original (pre-downsampled) visual local features through a residual connection, to restore spatial information. Finally, before passing to the decoder, TSalV360 retains only the last ERP frame from the residual-enhanced output, as the decoder predicts the saliency map for the last frame of each input sequence of ERP frames (similarly to \cite{salvit360}). This process results in $\{ \mathbf{Z_F}^{(m)} \in \mathbb{R}^{T \times C_m \times H_m \times W_m} \}_{m=1}^{M}$. 

\textbf{Hierarchical skip connections.} The decoder consists of five blocks: the first four comprise convolutional layers, followed by normalization, ReLU activation and upsampling layers; the last block includes only a convolutional layer followed by a sigmoid activation layer to produce the output saliency map. In addition, hierarchical skip connections have been introduced to apply concatenation with intermediate features along the channel dimension from earlier decoder blocks, allowing the model to preserve multi-scale spatial information. The output of the decoder consists of a set of saliency maps, one per tangent image, which are represented as $\mathbf{Y_{\text{projected}}} \in \mathbb{R}^{1 \times P_{\text{out}} \times P_{\text{out}} \times T}$. These maps are subjected into reverse tangent projection to form the final saliency map for the last ERP frame of the input sequence, $\mathbf{Y_{\text{output}}} \in \mathbb{R}^{1 \times H_{\text{out}} \times W_{\text{out}}}$.

\section{The TSV360 Dataset}
TSV360 contains videos up to 60 sec. long, from the VR-EyeTracking and Sports-360 datasets. Hence, its visual content spans a wide and diverse range, including e.g., indoor and outdoor scenes, sports events and short films. The fixation maps of the VR-EyeTracking and Sports-360 videos were obtained in \cite{8578657}, \cite{10.1007/978-3-030-01234-2_30} using an HTC VIVE Head-Mounted Display, capturing the head and gaze directions of $45$ participants, and the eye fixation points from $27$ participants, respectively.

\textbf{Ground-truth saliency maps.} First, we standardized the saliency annotations from the used datasets, by applying a Gaussian filter with a fixed standard deviation of $\sigma = 5^\circ$ to the raw fixation maps (following \cite{gaussianSigma}) to ensure consistency across all saliency maps. Then, as depicted in Fig.~\ref{fig2}, each ERP video and the corresponding saliency maps were processed by a fine-tuned version of the 2D video production algorithm from \cite{10.1007/978-3-031-53302-0_15}. As a reminder, this algorithm: i) detects salient regions through DBSCAN-based clustering using intensity and distance, ii) forms spatial-temporal sub-volumes across frames, iii) reduces abrupt visual changes by adding missing frames, iv) extracts fields of view (FOVs) around salient regions from the ERP frames, and v) stitches these 2D fragments in chronological order, forming 2D videos. To improve its efficiency, we: 
i) replaced DBSCAN with HDBSCAN \cite{hdbscan} that is better suited for variable-density clusters and does not require parameter tuning, ii) employed Haversine distance \cite{haversine} to define spatial-temporally correlated 2D sub-volumes, as it is better suited for preserving spatial relationships in $360^\circ$ scenes compared to Euclidean distance, and iii) applied a fine-tuned approach to generate distinct saliency maps for each individual salient event in the original ground-truth saliency map.

\textbf{Ground-truth textual descriptions.} Each of the obtained 2D videos was densely captioned (per second) using the SOTA LlaVA-Next-7B vision-language model \cite{li2024llavanext-strong}\footnote{Available at: https://huggingface.co/llava-hf/LLaVA-NeXT-Video-7B-hf} and the following prompt: ``Briefly describe what is depicted in the video, using one sentence''. Our goal was to extract a rich and varying set of captions for each 2D video, in order to train a model to focus on different salient objects/events in a video based on the provided textual description. Through this process, we saw that many descriptions referred to a single recurring event over the entire $360^\circ$ video, lacking semantic diversity and limiting the usefulness of these videos. So, we discarded around $49\%$ of the videos that did not contain multiple identifiable salient objects or events. In addition, we removed around $40\%$ of the generated pairs of saliency maps and textual descriptions that were not associated to any identifiable object or event. Finally, we manually curated around $65\%$ of the descriptions of the remaining data that were unclear or repetitive across events, to obtain more diverse and contextually relevant annotations.

\textbf{Data augmentation.} We observed that many of the selected $360^\circ$ videos contained an uneven distribution of events over time. In several cases, a single event was presented in a large part of the video while other events was shown only briefly. This imbalance resulted in significantly fewer training samples for the less frequent events, which posed a challenge for learning diverse text-grounded saliency detection patterns. To address this problem, we implemented a temporal window-shifting strategy during data sampling, based on the fact that our TSalV360 method takes as input a sequence of $F$ ERP frames and predicts the saliency map for the final one. In particular, we took multiple overlapping sequences of ERP frames in the case of shortly depicted events, increasing significantly the number of training samples for these events. Additionally, we further augmented the amount of obtained pairs of data, by producing paraphrased but semantically aligned text descriptions. The finally created TSV360 dataset comprises 160 $360^\circ$ videos, with approx. 16,000 triples of ERP frames, ground-truth saliency maps and textual descriptions.

We should note that the saliency information is obtained from fixations that are generic, not guided by a textual prompt. This is intentional; if the users had been prompted to look for specific objects, their attention could have focused on relevant but non-salient parts of the $360^\circ$ videos. We argue that the way we produce individual object/event-related ground-truth saliency maps and associate each of them with semantically-relevant textual descriptions, makes TSV360 suitable for training methods for text-driven saliency detection in $360^\circ$ videos.

\begin{table*}[]
\caption{The best scores are shown in bold. The arrows indicate the optimal (lower or higher) value for each evaluation measure.}
\resizebox{\textwidth}{!}{%
\begin{tabular}{|l|c|c|c|c|c|c|c|c|c|}
\hline
                     & \begin{tabular}[c]{@{}c@{}}Visual \\ features\end{tabular} & \begin{tabular}[c]{@{}c@{}}Text \\ features\end{tabular} & \begin{tabular}[c]{@{}c@{}}Skip \\ con.\end{tabular} & \begin{tabular}[c]{@{}c@{}}Similarity \\ estimation\end{tabular} & \begin{tabular}[c]{@{}c@{}}Attention\\ Mechanism\end{tabular} & Decoder                        & CC ($\uparrow$)                    & SIM ($\uparrow$)                    & KLD ($\downarrow$)                    \\ \hline
\begin{tabular}[c]{@{}l@{}}SalViT360 \cite{salvit360}\\ (Baseline)\end{tabular} & {\color[HTML]{21AAA5} ResNet-18}                            & \ding{55}                                                        & \ding{55}                                                    &\ding{55}                                                               & {\color[HTML]{21AAA5} VSTA}                                   & {\color[HTML]{21AAA5} ReLU}    & 0.382 ± 0.024          & 0.189 ± 0.013          & 18.486 ± 0.427          \\ \hline
Variant 1            & {\color[HTML]{21AAA5} ResNet-18}                            & {\color[HTML]{21AAA5} CLIP}                              & \ding{55}                                                          & \ding{55}                                                    & {\color[HTML]{EB5240} VSTCA}                          & {\color[HTML]{21AAA5} ReLU}    & 0.411 ± 0.019          & 0.260 ± 0.013           & 16.117 ± 0.410          \\ \hline
Variant 2            & {\color[HTML]{21AAA5} CLIP}                                & {\color[HTML]{21AAA5} CLIP}                              & \ding{55}                                                           & \ding{55}                                                    & {\color[HTML]{EB5240} VSTCA}                          & {\color[HTML]{EB5240} ReLU} & 0.472 ± 0.027          & 0.327 ± 0.025          & 14.434 ± 0.803          \\ \hline
Variant 3            & {\color[HTML]{21AAA5} CLIP}                                & {\color[HTML]{21AAA5} CLIP}                              & \ding{55}                                                          & \ding{55}                                                   & {\color[HTML]{EB5240} VSTCA}                          & {\color[HTML]{EB5240} Sigmoid} & 0.483 ± 0.029          & 0.346 ± 0.019          & 13.551 ± 0.672          \\ \hline
Variant 4            & {\color[HTML]{21AAA5} CLIP}                                & {\color[HTML]{21AAA5} CLIP}                              & \ding{55}                                                           & \ding{51}                                                  & {\color[HTML]{EB5240} VSTCA}                          & {\color[HTML]{EB5240} Sigmoid} & 0.535 ± 0.007          & 0.381 ± 0.011          & 12.241 ± 0.293          \\ \hline
\begin{tabular}[c]{@{}l@{}}TSalV360\\ (Proposed)\end{tabular}  & {\color[HTML]{21AAA5} CLIP}                                & {\color[HTML]{21AAA5} CLIP}                              & \ding{51}                                                  & \ding{51}                                                             & {\color[HTML]{EB5240} VSTCA}                                  & {\color[HTML]{EB5240} Sigmoid} & \textbf{0.541 ± 0.027} & \textbf{0.395 ± 0.018} & \textbf{11.718 ± 0.650} \\ \hline
\end{tabular}}
\label{tab:results}
\end{table*}

\section{Experiments}

\subsection{Evaluation protocol}

We evaluated the performance of TSalV360 on the TSV360 dataset using three evaluation measures from the literature, following \cite{metrics}, namely the Correlation Coefficient (CC), Similarity (SIM), and Kullback–Leibler Divergence (KLD). These measures were computed for each pair of machine-generated and ground-truth saliency map. To increase the number of experiments and the robustness of our evaluation, we split TSV360 into five non overlapping folds and performed 5-fold cross-validation using each time 80\% of the data for training and the remaining 20\% for testing. The reported experimental results represent the average score across the 5 folds.

\subsection{Implementation details}

Following \cite{salvit360}, videos were downsampled to 16 fps and sequences of $F = 8$ ERP frames at a resolution of $H_\text{in} \times W_\text{in} = 960 \times 1920$ were used as input to the model. The corresponding ground-truth saliency maps had a resolution of $480 \times 960$. The input frames were projected into $T = 18$ tangent images, 
that share the same resolution of $P_\text{in} \times P_\text{in} =  224 \times 224$ pixels and a FOV of $80^\circ$. Feature extraction was based on the CLIP model \cite{clip} with a ResNet-50 backbone as the visual encoder. The global features for both visual and textual data had a channel size of $C_G = 1024$, while the local textual features also use $C_L = 1024$. Local visual features were extracted at $M=3$ spatial scales $P_\text{in}/8$, $P_\text{in}/16$, and $P_\text{in}/32$, corresponding to channel sizes of $512$, $1024$ and $2048$, respectively. The local textual features had a token length of $L_t = 77$. VSTCA employed multi-head transformers with $8$ attention heads. The multi-layer perceptron (MLP) blocks in the transformer layers had a hidden dimension of $4 \times C_m$, for $m=1,...,M$. The decoder consisted of 2D conv. layers with kernel size $3 \times 3$ and stride $1$. The final decoder output had a shape of $P_\text{out}\times P_\text{out}= 56\times 56$ for each of the $T$ tangent views. This output was re-projected to $H_{\text{out}} \times W_{\text{out}}=240 \times 480$ and upsampled using bilinear interpolation to $480 \times 960$, to match the resolution of the ground-truth saliency maps for training and evaluation. Training was performed for $4$ epochs using a batch size of $8$ and the AdamW optimizer with learning rate equal to $1 \times 10^{-5}$. 
All experiments were conducted on an NVIDIA GeForce RTX 4090 GPU. The code for reproducing the reported results is publicly available at: \url{https://github.com/IDT-ITI/TSalV360}.

\begin{figure*}[t]
\centering
\includegraphics[width=0.82\textwidth]{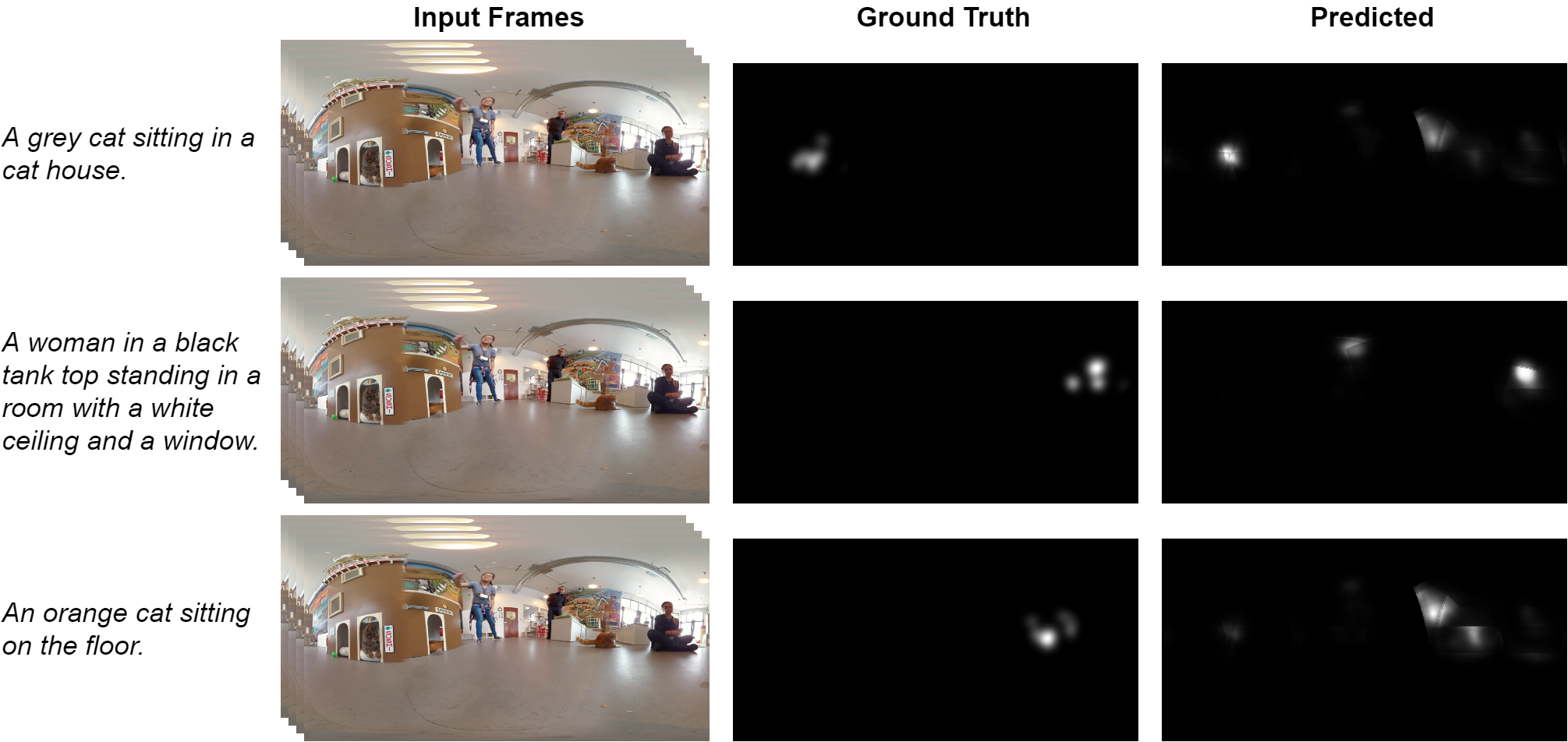}
\caption{Qualitative comparisons between the ground truth and the predicted saliency map generated by TSalV360 in an indoor scene.} 
\label{fig3}
\end{figure*}

\subsection{Quantitative results and ablations}

To evaluate the performance of the proposed TSalV360 method and assess the influence of the introduced changes in the visual-based SalViT360 method, we conducted an ablation study involving the following variants of TSalV360:
\begin{itemize}
    \item \textbf{Variant 1} extends SalViT360 by introducing the use of text as input and replacing VSTA with VSTCA.
    \item \textbf{Variant 2} replaces ResNet18 with CLIP's visual encoder and re-trains the decoder using the new visual features.
    \item \textbf{Variant 3} replaces ReLU with sigmoid in the decoder, to better align with the saliency prediction task.
    \item \textbf{Variant 4} introduces the SimEst module to take into account the relevance between ERP frames and text.
\end{itemize}

The results of this study are presented in Table \ref{tab:results}. The performance of Variant 1 indicates that the use of a textual description as proposed (i.e., using the VSTCA mechanism) leads to a clear improvement compared to SalViT360, pointing out the limited capacity of visual-based methods to perform customized saliency detection in $360^\circ$ videos. The results for Variant 2 show that using a vision-language model for representing the content from both modalities of the input data, and re-training the decoder with the new type of visual features, results in further performance gains. The outcomes for Variant 3 verify our initial assumption about the appropriateness of a sigmoid activation layer - which constrains the output to the expected saliency values (range [0, 1]) - since the replacement of ReLU led to further advancement of the saliency detection performance. The obtained scores for Variant 4 document the positive influence of the SimEst module, as the promotion of the frames that were more semantically relevant to the input text, resulted in noticeable gains in performance. Finally, the introduction of hierarchical skip connections to preserve multi-scale spatial information, also contributes positively according to all measures. These findings indicate that, through a set of well-designed and experimentally-validated changes in the SalViT360 method, the developed TSalV360 method can clearly better meet the needs of text-driven saliency detection in $360^\circ$ videos, establishing a strong baseline for future comparisons on the created TSV360 dataset. 

\subsection{Qualitative results}

Our qualitative analysis was based on manual observation of the generated saliency maps for several sequences of ERP frames and multiple textual descriptions per sequence. One of the examined samples is presented in Fig. \ref{fig3}, where a sequence of ERP frames is associated with different textual descriptions and the relevant pairs of ground-truth and predicted saliency maps from TSalV360. The sequence of frames shows an indoor scene with several objects and events taking place in parallel. So, the original ground-truth saliency map contains multiple salient regions that relate to these different objects and events. However, after taking into account each textual description, TSalV360 focuses only on the relevant regions of the ERP frames, producing saliency maps that are very close to the ground-truth ones. This example demonstrates also the level of semantic understanding and spatial granularity of TSalV360. Our method correctly makes the distinction between the grey cat inside the cat house and the orange cat on the floor. Such an observation demonstrates TSalV360's ability to appropriately combine visual and textual information for accurate saliency detection in $360^\circ$ videos.

\section{Conclusions}
In this paper, we presented a newly created dataset (TSV360) and a method (TSalV360) for text-driven saliency detection in $360^{\circ}$ video. The former comprises 16,000 triplets of ERP frames, textual descriptions and ground-truth saliency maps. The latter leverages a SOTA vision-language model for representing input data from different modalities and discovers dependencies among them using a similarity estimation module and a viewport spatio-temporal cross-attention mechanism. Experimental evaluations and ablations using the TSV360 dataset documented the contribution of various components of the TSalV360 method and indicated its capacity to perform $360^\circ$ video saliency detection conditioned to the input text.

\balance

\end{document}